%% file: icassp2022-multilingual-bbpe.tex
\documentclass{article}
\usepackage{spconf,amsmath,amssymb,multirow}

\input{new_cmds}

\ninept

\title{Bilingual End-to-end ASR with Byte-level Subwords}

\name{Liuhui Deng, Roger Hsiao, and Arnab Ghoshal}

\address{Apple \\
{\small \tt \{liuhui\_deng, rhsiao, aghoshal\}@apple.com}}

\begin{document}

\maketitle

\begin{abstract}
\input{abstract.tex}
\end{abstract}

\input{intro.tex}
\input{bbpe.tex}

\input{expr.tex}
\input{analysis.tex}

\input{conclusions.tex}

\input{acknowledgements.tex}

\bibliographystyle{IEEEbib}
\bibliography{bib/abbrev,bib/e2e}
\end{document}

%% file: new_cmds.tex
\newcommand{\beqn}{\begin{eqnarray}}
\newcommand{\eeqn}{\end{eqnarray}}
\newcommand{\beqnx}{\begin{eqnarray*}}
\newcommand{\eeqnx}{\end{eqnarray*}}

%% file: abstract.tex

In this paper, we investigate how the output representation of an end-to-end neural network affects multilingual automatic speech recognition (ASR).
We study different representations including character-level, byte-level, byte pair encoding (BPE), and byte-level byte pair encoding (BBPE) representations,
and analyze their strengths and weaknesses. We focus on developing a single end-to-end model to support utterance-based bilingual ASR, where speakers 
do not alternate between two languages in a single utterance but may change languages across utterances. We conduct our experiments on English 
and Mandarin dictation tasks, and we find that BBPE with penalty schemes can improve utterance-based bilingual ASR performance by 2\% to 5\% relative 
even with smaller number of outputs and fewer parameters. 
We conclude with analysis that indicates directions for further improving multilingual ASR.

%% file: intro.tex
\section{Introduction}
\label{sect:intro}

End-to-end (E2E) neural network based automatic speech recognition (ASR) provides a degree of flexibility and performance that makes E2E neural models
 an attractive option for multilingual ASR. A number of studies~\cite{kannan2019}~\cite{punjabi2020}~\cite{pratap2020} have focused on building a single E2E
model with multiple languages. The resulting E2E model can perform utterance-based multilingual ASR.
The works in~\cite{keli2019}~\cite{shan2019}~\cite{qiu2020}~\cite{punjabi2021} aim
to build an E2E model that can improve code switching.
While these approaches are different from each other, there are some similarities among them.
First, they aim to build a single E2E model to realize multilingual ASR. Second, the outputs of these models are often unions of the
characters or subwords of the target languages. One exception would be the work in~\cite{li2019}, which proposes to use
UTF-8 codewords, a byte-level representation, to represent the target languages.

Byte-level models have been proposed for natural language processing (NLP)~\cite{gillick2016}~\cite{ruiz2017byte}~\cite{xue2021}.
The idea is to convert text to a sequence of variable-length UTF-8 codewords, and to have the model predict one byte at each decoding step.
The advantages of byte-level representation are compactness and universality, as any combination of languages may be represented with an output dimension of only 256. 
However, a sequence represented at the byte level is always much longer than its character-level counterpart for languages such as Chinese and
Japanese \cite{wang2020neural}, which is because many characters of these languages are represented by multiple bytes in UTF-8.
As a result, a byte-level model can be error-prone since it needs to make multiple predictions for many single characters, 
and each prediction has a chance to make a mistake.
To compensate for this drawback, \cite{wang2020neural} proposes byte-level subwords for neural machine translation.
The idea is to apply byte pair encoding (BPE) \cite{sennrich2016neural} to UTF-8 codeword sequences and as a result,
an approach referred to as byte-level BPE (BBPE). BBPE inherits the advantages of UTF-8 byte-level representation. 
BBPE is able to represent all languages while keeping the output dimension in check. At the same time, as BBPE tokens are in general longer than 
byte-level tokens, the approach reduces the number of steps required by the decoding process.

In this work, we investigate bilingual (English and Mandarin) E2E ASR models by exploring different types of output representations,
including character-level, BPE, byte-level (UTF-8) and BBPE. Similar to some of the previous work cited, 
we build a single E2E model for utterance-based bilingual speech recognition.
Our contributions are threefold. First, we compare the strengths and weaknesses of different
output representations in monolingual and bilingual use cases. Second, we propose a method to adjust the bigram statistics in the BPE algorithm
and show that the BBPE representation leads to accuracy improvements in the bilingual scenario. Finally, we analyze different representations and show
how we might improve them for multilingual ASR.


%% file: bbpe.tex
\section{Output Representations for E2E ASR}
\subsection{Character-level Representation}

Using a character-level representation in an E2E model means that the output symbol set for the model is the set of graphemes of the target language.
In addition to graphemes, the output representation may also contain punctuation marks, digits, emojis or special tokens such as begin-of-sentence (BOS) or end-of-sentence (EOS).
According to \cite{zhou2018comparison} \cite{zou2018comparable}, character-level representation is often a good representation for Mandarin E2E models, and
this serves as one of the baselines in our experiments.

\subsection{BPE Representation}
The BPE algorithm~\cite{sennrich2016neural} starts from the character representation and iteratively merges the most frequent bigrams 
given a training text corpus.
At the end of this process, the BPE algorithm produces a symbol set that consists of subwords with different lengths. 
This symbol set can then be used by an E2E model as its output units.
It is common to keep the single characters in the final symbol set, so unseen words in the test set can still be represented by the symbol set.
For English, BPE is widely used in E2E ASR systems, as it improves accuracy and reduces computation due to the use of frequent subwords 
and the resulting shorter labeling sequences.


\subsection{Byte-level Representation}
\label{sect:bytes}

Scalability is one of the important aspects in designing an output representation for a multilingual E2E ASR model. As the model supports more languages, the
size of the symbol set increases. To tackle this problem \cite{li2019} proposes a byte-level representation based on UTF-8. Instead of using
characters or subwords as the symbols, byte-level model uses UTF-8 codewords as the output symbol set. The resulting representation is compact as each 
UTF-8 codeword only has 256 values so each symbol uses one byte. Yet, this representation is capable of representing any language, 
and adding more languages does not increase the size of the symbol set, which is an advantage compared to the character-level and BPE representation.
However, byte-level representation has two drawbacks, first, it increases the length of the sequence by up to 4x~\cite{wang2020neural}, and it
increases the number of decoding steps during inference. Second, not all byte sequences are valid UTF-8 sequences, which means the byte-level models may
generate invalid byte sequences that require special handling.

To repair an invalid byte sequence, \cite{li2019} proposes a dynamic programming algorithm to recover the Unicode characters given any byte sequence. We use this post-processing approach to recover characters from byte sequences as much as possible. 

\subsection{Byte-level BPE Representation}
\label{sect:penalty}

To circumvent the increase of sequence length for byte-level representation, \cite{wang2020neural} proposes byte-level BPE (BBPE) 
for neural machine translation, which applies BPE to byte-represented text. 
The advantage of this approach is that it reduces the sequence length by adopting frequent byte-level subwords and it
keeps the size of the symbol set in check. It is important to note that BBPE is equivalent to BPE for many Latin-based languages, since in UTF-8, all Latin
characters are single byte units. However, for languages like Chinese or Japanese, characters can use multiple bytes, so BBPE could be helpful.
Similar to BPE representation, BBPE representation might generate invalid byte sequences, and post-processing using dynamic programming is necessary 
to remedy that. Another aspect is that if we keep all the single-byte UTF-8 codewords in the symbol set after BPE, BBPE can represent all languages, 
as with the byte-level representation.

In this paper, we propose two penalty schemes to adjust the bigram statistics used by the BPE algorithm, and we only apply the penalty schemes to the Mandarin BBPE symbol extraction. 
The first one is length penalty. We define $LP_{b}$ as the length penalized number of occurrences for bigram $b$
\begin{eqnarray}
  LP_{b}(\alpha, l, c)=
\begin{cases}
c & l \leq N\\
(1-\alpha)c & l>N \label{eq:2}
\end{cases}
\end{eqnarray}
where $\alpha$ is the length penalty factor $(0 \leq \alpha \leq 1)$; $l$ is the length of bigram; $c$ is the bigram count
and $N$ is the cutoff point determining where to apply this penalty.

The purpose of length penalty is to penalize byte-level subwords longer than $N$ so as to encourage the BPE algorithm to form more short subwords.
For multibyte languages, many of these short subwords correspond to full characters, thus forming more short subwords leads to fewer subwords which are fractions of full characters.

The second penalty scheme is alphabet penalty, which penalizes alphabetic bigrams to suppress English subwords occurring in the Mandarin corpus.
The saved space will be distributed to Mandarin BBPE symbols. We define $AP_{b}$ as the alphabet penalized number of occurrences for bigram $b$
\begin{eqnarray}
AP_b(\beta, LP_b)=
\begin{cases}
(1-\beta) LP_b & \textsf{if b is alphabetic}\\
LP_b & \textsf{otherwise}
\end{cases}
\end{eqnarray}
where $\beta$ is the alphabet penalty factor $0 \leq \beta \leq 1$.
\label{sect:bbpe}

%% file: expr.tex
\section{Experimental Results}
We evaluate our approach through two sets of experiments on proprietary English and Mandarin dictation tasks.
First, we conduct the experiments on research purpose English and Mandarin data sets with 1k hours for each language, and hence,
the training data of the bilingual system is 2k hours in total (1k hours for each language).
Second, we train the baseline models and BBPE-based models on data sets with 5k hours for each language
to validate the findings. For the E2E models, we follow the procedure mentioned in~\cite{hsiao2020} to build our listen, 
attend and spell (LAS) models~\cite{chan2016}.
For the output representations involving BPE, we compute the BPE symbol sets on the corresponding training transcripts. 
For evaluation, our models are evaluated on 62 hours of English data and/or 35 hours of Mandarin data.

In this work, we assume the character set of the target language is known and there is no unseen character during evaluation.
As a result, the symbol set of English to be used in our experiments consists of 765 symbols, including English alphabets, punctuation marks, digits, emojis and several hundreds of Unicode characters that appear in our in-house training datasets, such as cent sign. For Mandarin, there are 7632 symbols. 
In the experiments with byte-level (UTF-8) symbol set, we add 6 additional special tokens, including BOS and EOS, resulting in 262 symbols in total. 



\begin{table*}[tbp]
\centering
\caption{WER/CERs of the mono-/multilingual E2E Model using different types of output representations, 1k hrs experiments}
\vspace{1.0em}
\label{table:1k_expr}
\begin{tabular}{c|c |c |c | c| c|c} 
 \hline
Model & Exp. & Output Rep. & Output Dim. & Params & En & Zh  \\ [0.5ex] 
 \hline
 \multirow{3}*{Mono. En} & M0  & BPE & 6917  & 75M & \textbf{11.6} & -\\ 
 & M1 &character  &   765 & 50M & 13.8 &-\\ 
 &  M2  & UTF-8 & 262 & 48M & 13.6 &-\\ 
 \hline
 \multirow{6}*{Mono. Zh} &M3 &  character & 7632 & 78M& - &14.6 \\ 
  & M4 & UTF-8& 262 & 48M & - & 17.8\\ 
  &   M5 & BBPE& 3658 & 62M & - & 15.7\\ 
 & M6 & BBPE + LP(0.6) & 3662 & 62M & - & 15.1 \\ 
  & M7 & BBPE + LP(0.99) & 3661 & 62M & - & 14.8 \\ 
 & M8 & BBPE + LP(0.99) + AP(0.999) & 3655 & 62M & - & \textbf{14.5} \\ 
 \hline
 \multirow{5}*{Bi. En + Zh} & B0 & BPE(En) + character(Zh) & 14414 & 105M & 12.1 & 14.6\\ 
 & B1 & character & 8115 & 80M &13.5 &14.9 \\ 
  & B2 & UTF-8& 262 & 48M & 14.0 & 18.1\\ 
  & B3 & BBPE & 7028 & 75M & 11.5 & 15.3\\ 
 & B4 & BBPE + LP(0.99) + AP(0.999) & 7140 & 76M & \textbf{11.5} & \textbf{14.3} \\ [0.5ex] 
 \hline
\end{tabular}
\end{table*}

\begin{table*}[tbp]
\centering
\caption{WER/CERs of the best BBPE-based E2E Models and baselines, 5k hrs experiments}
\vspace{1.0em}
\label{table:5k_expr}
\begin{tabular}{c|c |c |c | c| c|c}
 \hline
 Model & Exp. & Output Rep. & Output Dim. & Params & En & Zh  \\ [0.5ex] 
 \hline
Mono. En & M9  &  BPE & 7091 &75M & \textbf{6.4} & -\\ 
 \hline
  \multirow{3}*{Mono. Zh} &M10 & character & 7632 & 78M & - &\textbf{9.3} \\ 
    &M11 & BBPE&  3676 & 62M & - & 9.9\\ 
 &  M12 & BBPE + LP(0.99) + AP(0.999) &3674 & 62M& - & 9.4 \\ 
 \hline
  \multirow{3}*{Bi. En + Zh} & B5 &BPE(En) + character(Zh) & 14577 & 105M & 7.2 & 10.2\\ 
  &B6 & BBPE &  7057 & 75M& 7.2 & 10.6\\ 
&  B7  & BBPE + LP(0.99) + AP(0.999) &7170 & 76M& \textbf{7.0} & \textbf{9.9} \\ [0.5ex] 
 \hline
\end{tabular}

\end{table*}

\label{sect:expr}
\subsection{Monolingual models}
In the English monolingual experiments, M0, M1 and M2 in Table~\ref{table:1k_expr} correspond to the English LAS 1k hours models using BPE,
character-level and byte-level representation respectively. The system using BPE representation achieves 11.6\% word error rate (WER),
which outperforms character-level and byte-level representation by 15\% relative.
It is important to note that character-level and byte-level representations are similar for English except that the
character-level representation contains special tokens like emojis.

For Mandarin monolingual results, since Chinese characters use multiple bytes in UTF-8, there is a big accuracy gap
between character-level (M3) and byte-level (M4) representations. The character-level system has a character error rate (CER) of 14.6\%,
where the byte-level representation is worse by 3.2\% absolute or 21.9\% relative.
Applying BPE on the byte-level, i.e. BBPE (M5), recovers some of the degradation but the gap to the character-level system is still 1.1\% absolute.
We observe that the difference in accuracy between the BBPE system and the character-level system is mostly due to deletion errors.
These deletion errors are caused by invalid UTF-8 codeword sequences generated by the model. 
Although the dynamic programming algorithm only produces valid sequences, it cannot recover the correct bytes in many cases.

By using the length penalty discussed in Section~\ref{sect:penalty} (M7), we can shrink the gap to 0.2\% absolute with a length penalty factor of 0.99.
For the length penalty, we choose the cutoff $N=3$ as most of the Chinese characters consist of three bytes. 
The penalty scheme discourages the BPE algorithm from generating multi-character symbols, encourages generating more single-character symbols, and reduces the chance of producing invalid byte sequences.
As for the length penalty factor $\alpha$, we tried values ranging from 0.6 to 0.99, as shown in Table~\ref{table:1k_expr}, M7 achieves the best CER. 

By further applying the alphabet penalty (M8), we find that the BBPE system can be as good as the character-level system even with 
smaller number of outputs and fewer parameters, as the alphabet penalty forces the BPE algorithm to suppress multibyte English symbols.
In this work, we choose the alphabet penalty factor $\beta=0.999$ (M8), with which there are only two multibyte English symbols in the symbol set, 
while in M7, 10\% of the symbols are multibyte English symbols. 
When scaled to the 5k hours training set, the conclusion remains the same: the BBPE representation with length and alphabet penalty 
can recover the degradation (M12 in Table~\ref{table:5k_expr}).

\subsection{Bilingual models}



The last five rows of Table~\ref{table:1k_expr} are the results of our 1k hours bilingual systems (B0 to B4). B0 is our baseline bilingual system where
it combines the BPE symbol set from the monolingual English system (M0), and the character-level symbol set from the monolingual Mandarin system (M3).
Compared to the corresponding monolingual systems, 
we observe 0.5\% absolute degradation on English and similar accuracy on Mandarin.
As expected, B0 has better accuracy on English than the character-level bilingual system (B1). The byte-level system (B2) shows significant
degradation, 1.9\% absolute on English and 3.5\% absolute on Mandarin, as seen in the monolingual experiments. 

Using BBPE representation recovers most of the degradation observed in B2.
In fact, the BBPE system (B3) is better than the baseline on the English test set by 0.6\% absolute, while there is still a 0.7\% gap on the Mandarin test set.
Finally, by applying the length and alphabet penalty, our best BBPE system (B4) outperforms the baseline (B0) by 0.6\% absolute on English 
and 0.3\% absolute on Mandarin. Besides, compared to the monolingual baselines (M0 and M3), B4 also shows slight improvement by 0.1\% absolute and 0.3\% absolute, respectively. 

When scaled to the 5k hours training set, the results remain consistent. While the BBPE system (B6) shows some degradation in accuracy 
when compared to the baseline (B5), using length and alphabet penalty recovers the loss and the best BBPE system (B7) shows slight improvement, 3\% relative on both English and Mandarin test sets. 
The only difference compared to the 1k hours experiments is that the 5k hours bilingual systems show small degradation when compared to their respective 
monolingual systems. One possible explanation is that as the amount of data increases, the bilingual systems might require a larger model.

%% file: analysis.tex
\section{Analysis}

\subsection{Invalid byte sequences and the effect of penalty mechanisms}

We notice that the BBPE representation may increase the deletion rate  when the model generates invalid byte sequences.
In Table~\ref{table:err}, when we compare the BBPE system (B6) with the baseline (B5), we can see a 19\% increase in deletions. 
However, the penalty mechanisms fully recover the increase in deletion errors (B7). 
Comparing the symbol sets of the BBPE-based bilingual experiments B6 (w/o penalties) and B7 (w/ penalties), we find that in B6 only 24\% of the symbols 
represent complete Mandarin characters and 20\% of them represent Mandarin multi-character sequences.
In B7, on the other hand, 42\% of the symbols represent complete Mandarin characters and less than 2\% represent Mandarin multi-character sequences.


\begin{table}[t]
\centering
\caption{Penalty mechanisms mitigate the increase of deletion errors on the Mandarin test set}
\vspace{0.5em}
\label{table:err}
\begin{tabular}{c c|c | c | c } 
 \hline
 Exp. & penalties & \#del & \#sub & \#ins \\ [0.5ex] 
 \hline
 B5&-& 2273 & 19176 & 804 \\ 
 B6 & No& 2700 (+19\%) & 19609 (+2\%)& 774 (-4\%) \\ 
 B7 & Yes &2164 (-5\%) & 18862 (-2\%)& 711 (-12\%)\\ [0.5ex] 
 \hline
\end{tabular}
\end{table}

\subsection{Symbol sharing across languages}
One motivation for using BBPE symbols is to allow more symbols to be shared in the multilingual scenario.
Symbol sharing rate of a bilingual model is measured based on symbol sets, it is defined as the ratio of symbols existing in both monolingual symbol sets to the number of symbols in the combined bilingual symbol set, which is simply a combination of the two monolingual symbol sets.
As shown in Table \ref{table:share}, in the baseline bilingual experiment B5, only 1\% of the symbols are shared.
In our BBPE bilingual experiment B7,
2.6\% of the symbols are shared between English and Mandarin.  Higher length penalty factor alone may lead to higher sharing rate, 
since multi-character Chinese symbols are suppressed and thus there are more multibyte English symbols in the Mandarin symbol set. 
But the sharing rate in B7 is not as high since higher alphabet penalty factor leads to lower sharing rate.
We expect alternative byte-level representations, as well as the choice of languages, can lead to higher sharing rates. 

\begin{table}[htbp]
\centering
\caption{Symbol sharing between English and Mandarin in bilingual symbol sets}
\vspace{0.5em}
\label{table:share}
\begin{tabular}{c|c | c } 
 \hline
 Exp. & \#total symbols & \#shared symbols \\ [0.5ex] 
 \hline
B5& 14577& 146 (1.0\%) \\ 
 B7 & 7170 & 186 (2.6\%) \\ [0.5ex] 
 \hline
\end{tabular}
\end{table}
\subsection{Language confusion}
In our experiments, no external language information is used in the bilingual models,
and the bilingual models need to be able to identify the language in the audio.
There is a possibility that the model would be confused between the two languages. For example,
an English utterance could be recognized as a Mandarin utterance that has similar pronunciation.
We investigate whether the output representation affects the confusion rate.

Table \ref{table:lang_c} shows the percentages of utterances that are recognized as the wrong language in the bilingual experiments B5 and B7.
We can see that the language confusion ratios of the two bilingual models are close,
around 0.2\% of the English utterances are recognized as Mandarin while 0.8\% of the Mandarin utterances are recognized as English.
The higher confusion rate observed in Mandarin test set can be explained by the higher occurrence of English words in the Mandarin training set, 
however, most of the confusions in Mandarin test set do not come from code-switched utterances but short utterances (1-2 characters).

Generally speaking, it seems the choice of output representation does not impact the confusion rate, which might be due to the low sharing rate
of the symbols between the two languages. From the perspective of the bilingual models, both character-level representation and 
byte-level representation provide two mostly mutually exclusive sets of symbols, one for each language. This could be the reason why
the confusion rates are similar.
To further reduce the confusion rate, we may consider various language identification approaches~\cite{kannan2019},~\cite{punjabi2020},~\cite{watanabe2017}.

\begin{table}[htbp]
\centering
\caption{Percentages of utterances that are recognized as the wrong language in bilingual models}
\vspace{0.5em}
\label{table:lang_c}
\begin{tabular}{c|c | c } 
 \hline
 Exp. & En recognized as Zh &Zh recognized as En \\ [0.5ex] 
 \hline
B5& 0.17\% &  0.80\%\\ 
 B7 & 0.21\% & 0.80\% \\ [0.5ex] 
 \hline
\end{tabular}
\end{table}

\subsection{The average length of hypotheses from BBPE-based bilingual models}
We calculate the average length of hypotheses from the bilingual experiments.
The length of a hypothesis is defined as the number of symbols.
We evaluate this metric as hypothesis length corresponds to the number of decoding steps required for recognition, which directly affects computation time. 
Hence, we would like to  measure the length of hypotheses under different output representations.

Table \ref{table:length} shows the average length of hypotheses from five bilingual models. 
Byte-represented hypotheses are much longer in both test sets, as shown in B2, which might explain why it suffers from significant accuracy degradation. 
On the English test set, the best-performing BBPE bilingual model B4, outputs longer hypotheses than the bilingual baseline B0, which is expected, 
since the English BBPE symbols of B4 are on average shorter than those of B0.
On the Mandarin test set, however, the hypotheses from B4 are shorter than B0, which we attribute to the use of multi-character symbols. 
The average length of the hypotheses from B4 on the Mandarin test set is greater than that for B3 due to length penalty.
\begin{table}[htbp]
\centering
\caption{Average length of hypotheses from bilingual models}
\vspace{0.5em}
\label{table:length} 
\begin{tabular}{c c|c | c|c } 
 \hline
 Exp. & Output Rep. &Output Dim.  & En  & Zh  \\ [0.5ex] 
 \hline
 B0 &BPE(En)+ character(Zh) &14414 & 26.3 & 9.6 \\ 
 B1 & character&8115 & 61.3& 9.7\\ 
 B2 & UTF-8&262 & 61.3& 25.7\\ 
 B3 & BBPE&7028 & 27.1& 7.7\\ 
 B4 & BBPE(w/ penalties)&7140 &27.2 & 8.8 \\ [0.5ex] 
 \hline
\end{tabular}
\end{table}
\label{sect:analysis}

%% file: conclusions.tex
\section{Conclusions}

In this paper, we compared different output representations for bilingual E2E ASR, including character-level, BPE, byte-level and BBPE representations.
We found that BBPE representation may cause higher deletion rate due to invalid byte sequences. To tackle that, we proposed
penalty mechanisms and the resulting BBPE-based bilingual system is shown to outperform the baseline bilingual system using a mixture of BPE and 
character-level representation. However, our 5k hrs BBPE-based bilingual system still lags behind the monolingual counterparts, we will try to increase the 
model capacity in an attempt to close the gap in future work. In our analysis, we noticed that the current BBPE representation has low sharing rate 
between the two languages 
which may be due to the nature of UTF-8 and in the future, we would look into alternative byte-level representations, and we believe it might lead to 
better bilingual performance.

\label{sect:conclude}

%% file: acknowledgements.tex
\section{Acknowledgments}
\label{sect:acknowledge}

We would like to thank Erik McDermott, Pawel Swietojanski, Russ Webb and Manhung Siu for their support and useful discussions.